\pdfoutput=1
\documentclass[11pt]{article}

\usepackage[hyperref]{naacl2021}
\usepackage{soul}
\usepackage{times}
\usepackage{latexsym}
\usepackage[T1]{fontenc}
\usepackage[utf8]{inputenc}
\usepackage{microtype}

\usepackage{amsmath}
\usepackage{amssymb} 
\usepackage{latexsym}
\usepackage{graphicx}
\usepackage{hyperref}
\usepackage{booktabs}
\usepackage{multirow}
\usepackage{url}
\usepackage{caption}

\usepackage{todonotes} 

\title{Multi-Adversarial Learning for Cross-Lingual Word Embeddings}

\author{
	Haozhou Wang\textsuperscript{\rm 1}, James Henderson\textsuperscript{\rm 2}, Paola Merlo\textsuperscript{\rm 1} \\
	\textsuperscript{\rm 1} University of Geneva\\
	\textsuperscript{\rm 2} Idiap Research Institute\\
	\{haozhou.wang, paola.merlo\}@unige.ch, james.henderson@idiap.ch
}

\begin{document}
\maketitle

\begin{abstract}

Generative adversarial networks (GANs) have succeeded  in inducing  cross-lingual word embeddings ---maps of matching words across languages--- without supervision. Despite these successes, GANs' performance for the difficult case of distant languages is still not satisfactory. These limitations have been explained by GANs' incorrect assumption that source and target embedding spaces are related by a single linear mapping and are approximately isomorphic. We assume instead that, especially across distant languages, the mapping is only piece-wise linear, and propose a multi-adversarial learning method. This novel method induces the seed cross-lingual dictionary through multiple mappings, each induced to fit the mapping for one subspace. Our experiments on unsupervised bilingual lexicon induction and cross-lingual document classification show that this method improves performance over  previous single-mapping methods, especially for distant languages.

\end{abstract}

\section{Introduction and background}
\label{section_introduction}

Word embeddings, continuous vectorial representations of words, have become a fundamental initial step in many natural language processing (NLP) tasks for many languages. In recent years,  their cross-lingual counterpart, cross-lingual word embeddings (CLWE) ---maps of matching words across languages--- have been shown to be useful in many important cross-lingual transfer and modeling tasks such as machine translation,  cross-lingual document classification and zero-shot dependency parsing \cite{klementiev2012inducingcrosslingual, zou2013bilingualword, guo2015crosslingualdependency, conneau2018wordtranslation, glavas2019howto, mozhi2020whyoverfitting}.

In these representations, matching words across different languages are represented by similar vectors. Following the observation of \citet{mikolov2013efficientestimation} that the geometric positions of similar words in two embedding spaces of different languages appear to be related by a linear relation, the most common method aims to map between two pretrained monolingual embedding spaces by learning a single linear transformation matrix. Due to its simple structure design and competitive performance, this approach has become the mainstream of learning CLWE \cite{glavas2019howto, vulic2019dowe, ruder2019asurvey}. 

Initially, the linear mapping was learned by minimizing the distances between the source and target words in a seed dictionary. Early work from \citet{mikolov2013efficientestimation} uses a seed dictionary of five-thousand word pairs. Since then, the size of the seed dictionary has been gradually reduced, from several-thousand to fifty word pairs \cite{smith2017offlinebilingual}, reaching a minimal version of only sharing numerals \cite{artetxe2017learningbilingual}.

More recent works on unsupervised learning  have shown that mappings across embedding spaces can also be learned without any bilingual evidence \cite{barone2016towardscrosslingual, zhang2017adversarialtraining, conneau2018wordtranslation, hoshen2018aniterative, alvarezmelis2018gromovwassersteinalignment, artetxe2018arobust}. More concretely, these fully unsupervised methods usually consist of two main steps \cite{hartmann2019comparingunsupervised}: an unsupervised step which aims to induce the seed dictionary by matching the source and target distributions, and then a pseudo-supervised refinement step based on this seed dictionary.

The system proposed by \citet{conneau2018wordtranslation} can be considered  the first successful unsupervised system for learning CLWE. They first use generative adversarial networks (GANs) to learn a single linear mapping to induce the seed dictionary, followed by the Procrustes Analysis \cite{Schoneman1966ageneralized} to refine the linear mapping based on the induced seed dictionary. While this GAN-based model has competitive or even better performance compared to supervised methods on typologically-similar language pairs, it often exhibits poor performance on typologically-distant language pairs, pairs of languages that differ drastically in word forms, morphology, word order and other properties that determine how similar the lexicon of a language is. More specifically, their initial linear mapping often fails to induce the seed dictionary for distant language pairs \cite{vulic2019dowe}. Later work from \citet{artetxe2018arobust} has proposed an unsupervised self-learning framework to make the unsupervised CLWE learning more robust. Their system uses similarity distribution matching to induce the seed dictionary and stochastic dictionary induction to refine the mapping iteratively. The final CLWE learned by their system  performs better than the GAN-based system. However, their advantage appears to come from the iterative refinement with stochastic dictionary induction, according to \citet{hartmann2019comparingunsupervised}. If we only consider the performance of a model induced only with distribution matching, GAN-based models perform much better. This brings us to our first conclusions, that a GAN-based model is preferable for seed dictionary induction.

Fully unsupervised mapping-based methods to learn CLWE rely on the strong assumption that monolingual word embedding spaces are isomorphic or near-isomorphic, but this assumption is not fulfilled in practice, especially for distant language pairs \cite{sogaard2018onthe}.  Experiments by \citet{vulic2020areall} also demonstrate that the lack of isomorphism does not arise only because of the typological distance among languages, but it also depends on the quality of the monolingual embedding space. If we replace the seed dictionary learned by an unsupervised distribution matching method with a pretrained dictionary, keeping constant the refinement technique, the final system becomes more robust \cite{vulic2019dowe}. 

All these previous results indicate that learning a better seed dictionary is a crucial step to improve unsupervised cross-lingual word embedding induction and reduce the gap between unsupervised methods and supervised methods, and that GAN-based methods hold the most promise to achieve this goal.
The results also indicate that a solution that can handle the full complexity of  induction of cross-lingual word embeddings will  show improvements in both close and distant languages.

In this paper, we focus on improving the initial step of distribution matching, using GANs  \cite{hartmann2019comparingunsupervised}. Because the isomorphism assumption is not observed in reality, we argue that a successful GAN-based model must not learn only one single linear mapping for the entire distribution, but must be able to identify mapping subspaces and learn multiple mappings. We propose a multi-adversarial learning method which learns different linear maps for different subspaces of word embeddings.

\section{Limitations of single-linear mappings}

\begin{figure}
\centering
\includegraphics[scale=0.26]{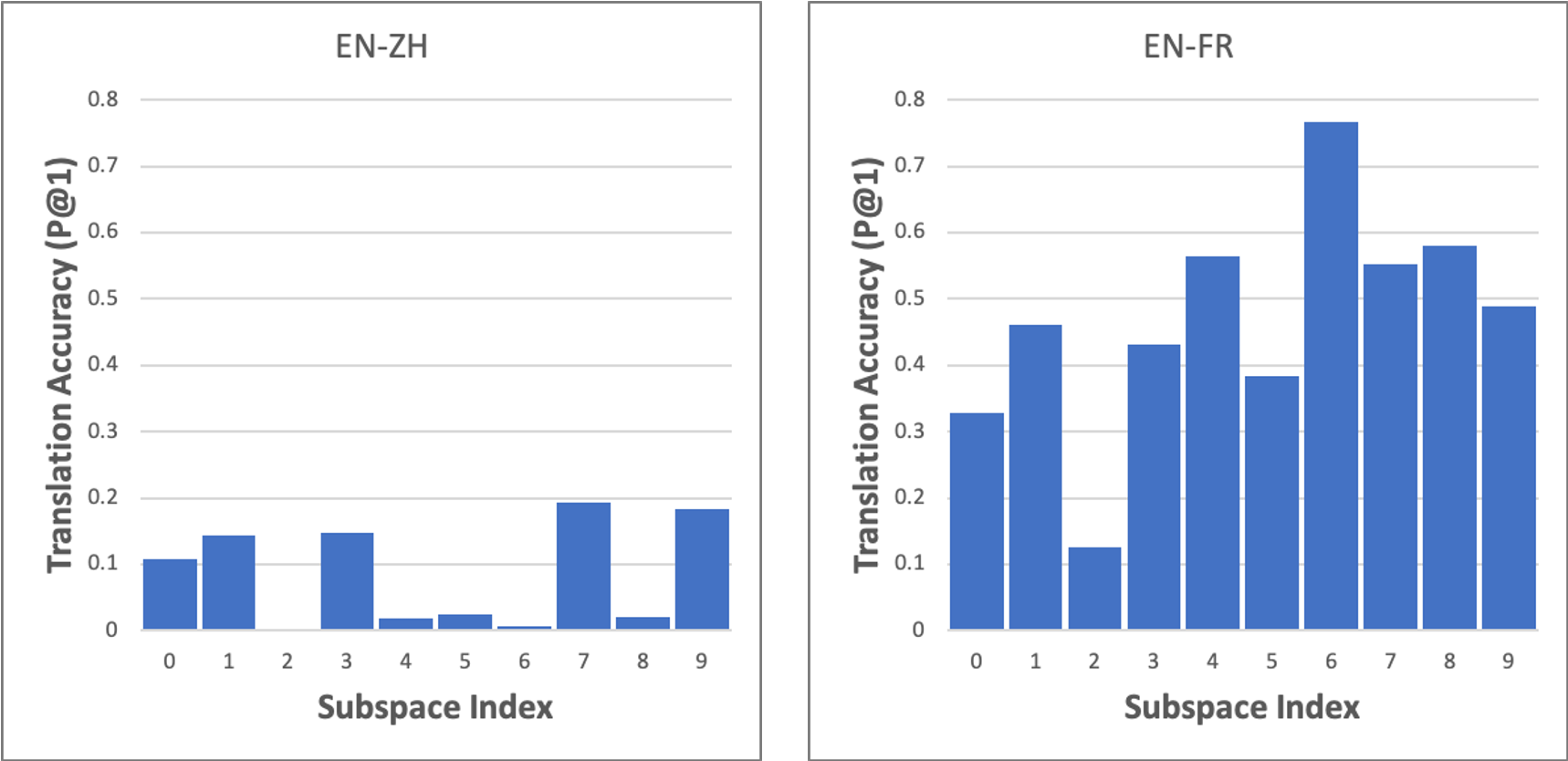}
\caption{Translation accuracy from English to Chinese and to French for different English subspaces. We only include the top fifty-thousand most frequent English words in the pretrained fastText embeddings. The gold translations comes from Google Translate.}
\label{figure_translation_accuracy_per_cluster}
\end{figure}

If the assumption by \citet{mikolov2013efficientestimation} that similar words across source and target languages are related by a single linear relation  holds exactly or even approximately , the distance between source and target embedding spaces should be (nearly) evenly minimized during the training of the initial mapping. More specifically, each source subspace should be mapped (nearly) equally well to its corresponding target space, so that the translation ability of the single linear mapping should be similar across different source subspaces. 

To verify this expectation, we use the GAN-based system MUSE\footnote{https://github.com/facebookresearch/MUSE}  to train two linear mappings (without refinement) \citep{conneau2018wordtranslation}. One mapping relates two typologically distant languages, English and Chinese, and the other maps the English space to the space of French - a typologically similar language. We use pretrained FastText embeddings.\footnote{https://fasttext.cc/docs/en/pretrained-vectors.html} 
We split the English space into ten subspaces by running $k$-means clustering. We evaluate the trained linear mappings by calculating the translation accuracy with precision at one (P@1) ---how often the highest ranked translation is the correct one--- for each subspace, using  the translations from Google Translate as the gold dataset. To reduce the influence of infrequent words, we only consider the first fifty-thousand most frequent source words.

As we can see in Figure \ref{figure_translation_accuracy_per_cluster}, the distribution of accuracies of different subspaces is not uniform or even nearly so.  This is true for both  language pairs, but particularly for the distant languages, where the general mapping does not work at all in some subspaces. Similar phenomena were also discovered by \citet{nadapa2018normaneighborhood} where source words are grouped into different categories. This lack of uniformity in results corroborates the appropriateness of designing a model that learns different linear mappings for different subspaces instead of only learning a single linear mapping for the entire source space.

\section{Multi-adversarial CLWE learning}
\label{section_learning_multi-linear_mapping_through_distribution_matching}

To learn different mappings for different source subspaces, we propose a method for training one GAN for each source subspace.  These multi-discriminator GANs encourage the distribution of mapped word embeddings from a specific source subspace to match the distribution of word embeddings from the corresponding target subspace.

The first step of our proposed method is to train a single linear mapping, as in previous approaches.  This is used 
to find aligned subspaces.  Our proposed multi-discriminator GAN model then learns the multi-linear mapping.
This section starts with the two GAN models, followed by the subspace alignment method, and then describes methods used to improve the GAN training.

\subsection{Unsupervised CLWE learning}
\label{section_unsupervised_clwe_learning}

We first define the task of learning CLWEs and the role of GANs in the previous work of \citet{conneau2018wordtranslation}.
Let two monolingual word embeddings $V_s^j=\{v_{s_{_1}}, ..., v_{s_{_j}}\}$ and $V_t^k=\{v_{t_{_1}}, ..., v_{t_{_k}}\}$ be given. In previous work, mapping $V_s^j$ to $V_t^k$  means seeking a linear transformation matrix $W$, so that the projected vector $Wv_i$ of a source word is close to the vector of its translation in the target language. The basic idea underlying supervised methods is using a seed dictionary of $n$ word pairs $\{(w_{s_{_1}}, w_{t_{_1}}), ..., (w_{s_{_n}}, w_{t_{_n}})\}$ to learn the matrix $W$ by minimizing the  distance in (\ref{equation_mapping_mikolov}), where $v_{s_{_i}}$ and $v_{t_{_i}}$ represent the embeddings of $w_{s_{_i}}$ and $w_{t_{_i}}$. The trained matrix $W$ can then be used to map the source word embeddings to the target space.
\begin{equation}
\label{equation_mapping_mikolov}
\min_{W}\sum_{i=1}^{n}\| Wv_{s_{_i}} - v_{t_{_i}} \|^{2}
\end{equation}

In an unsupervised setting, the seed dictionary is not provided. \citet{conneau2018wordtranslation} propose a two-step system where the seed dictionary is learned in an unsupervised fashion. In a first step, they use GANs to learn an initial linear transformation matrix $W$ and use this to induce a seed dictionary by finding the translations of the first ten-thousand most frequent source words. In a second step, the seed dictionary is used to refine the initial matrix $W$.  In this work we focus on the GAN component of this model.

\subsection{GAN learning of a single linear mapping}
\label{section_single_linear_mapping}

Previous  GAN-based systems learn a single linear mapping from the source embedding space to the target embedding space. In such models, a source word is trained against a target word sampled from the whole target distribution, and the resulting single linear mapping is applied to all the source words.
We first introduce the basic GAN architecture for CLWE of \citet{conneau2018wordtranslation}.
We use this model as our comparative baseline and as the initial stage of our proposed method.

A standard GAN model plays a min-max game between a generator $G$ and a discriminator $D$ \citep{goodfellow2014generativeadversarial}. The generator learns from the distribution of source data and tries to fool the discriminator by generating new samples which are similar to the target data. 

When we adapt the basic GAN model to learning CLWE, the goal of the generator is to learn the linear mapping matrix $W$.
The discriminator $D$ detects whether the input is from the distribution of target embeddings $p_{v_t}$.  \citet{conneau2018wordtranslation} use the loss functions in (\ref{equation_discriminator_loss}) and (\ref{equation_generator_loss}) to update the discriminator and the generator, respectively. $G(v_s) = Wv_s$, and $D(v)$ denotes the probability that the input vector $v$ came from the target distribution $p_{v_{t}}$ rather than the generator applied to samples from the source distribution $p_{v_{s}}$. 
\begin{align}
\label{equation_discriminator_loss}
l_D &= -\log D(v_t) - \log (1 - D(G(v_s)))
\\
\label{equation_generator_loss}
l_G &= -\log D(G(v_s)) - \log(1 - D(v_t))
\end{align}
The parameters of both generator and discriminator are updated alternatively by using stochastic gradient descent.
%
However, a number of additional methods are needed for robust reliable training of such GANs,
which are discussed in Section~\ref{section_training}.

\begin{figure}
\centering
\includegraphics[width=\columnwidth]{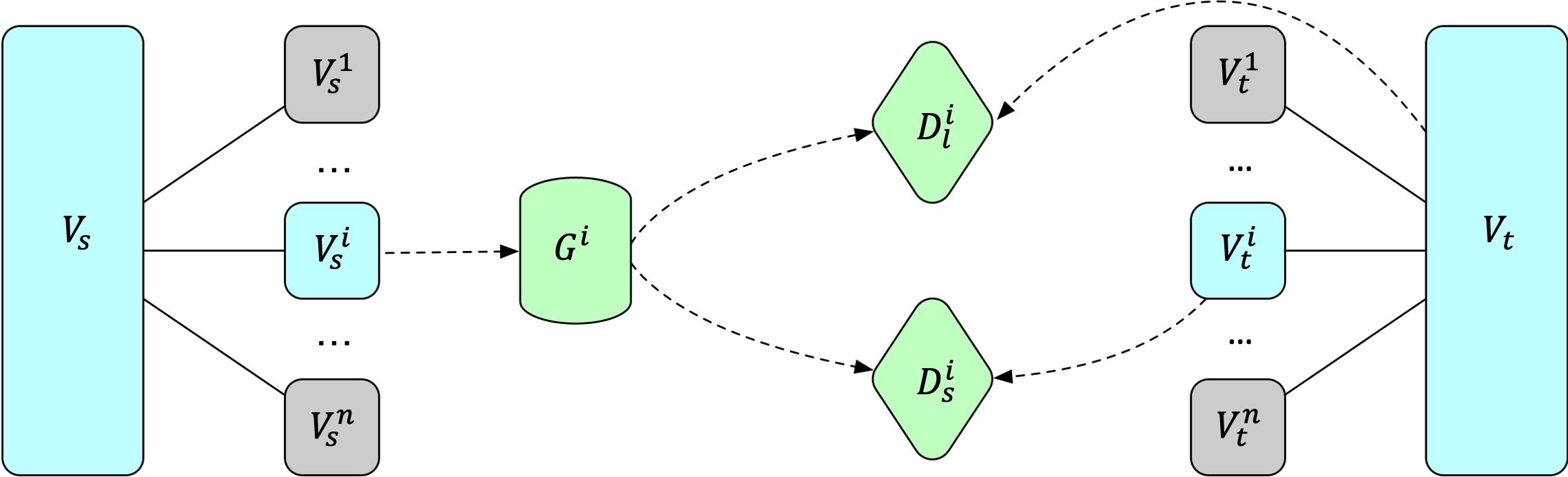} 
\caption{Architecture of our multi-discriminator model. The generator $G^i$ for each source subspace $V^i_s$ is trained against the discriminator $D^i_s$ for the aligned subspace $V^i_t$ and a whole-language discriminator $D^i_l$.
}
\label{figure_multi_discriminator_mapping}
\end{figure}

\subsection{GAN learning of a multi-linear mapping}
\label{section_multi_linear_mapping}

Unlike previous work, we propose learning different linear mappings for different source subspaces.  We propose a multi-discriminator GAN where a source word from one subspace is trained against a target word sampled from the aligned target subspace.

For each subspace of source embeddings, we propose a multi-discriminator adversarial model to train the specific mapping for vectors that belong to this subspace. As the architecture in Figure~\ref{figure_multi_discriminator_mapping} illustrates, the generator of the given source subspace $i$ takes the vector sampled from the sub-distribution as input and maps it to the target language. Differently from standard GANs, the mapped vector
$G^i(v^i_s){=}W^iv^i_s$
will be fed into two discriminators.
%
First, a subspace-specific discriminator $D^i_s$  judges whether the vector has come from the correspondent target subspace $i$. Thus, we use the vectors sampled from both source and target subspaces to train $D^i_s$.
Second, a normal language discriminator $D^i_l$ judges whether the vector has come from the whole target distribution.  This language discriminator helps avoid local optima for the specific subspace.

Both discriminators are two-layer perceptron classifiers. Except for the different sampling ranges, their loss function is similar to equations \eqref{equation_discriminator_loss} and \eqref{equation_generator_loss}:
\begin{align}
\label{equation_language_discriminator_loss}
l_{D^i_l} &= -\log D(v'_t) - \log (1 - D(G(v'_s)))
\\
\label{equation_subspace_discriminator_loss}
l_{D^i_s} &= -\log D(v^i_t) - \log (1 - D(G(v^i_s)))
\end{align}
where $v'_s$ and $v'_t$ are sampled from the 75-thousand most frequent source and target words,\footnote{We use different language discriminator models $D^i_l$ for each subspace $i$, even though their training samples all come from the same distributions.  This leads to more stable training, presumably because initially these language discriminators are randomly different.} and $v^i_s$ and $v^i_t$ are sampled from the specific source subspace $V^i_s$ and its corresponding target subspace $V^i_t$. Since the outputs of both discriminators are used for training the generator, the loss function of the subspace-specific generator $G^i$ can be written as:
%
\begin{equation}
\label{equation_subspace_specific_generator_loss}
\begin{split}
&l_{G^i} = -\lambda (\log D^i_l(G^i(v^i_s)) + \log(1 - D^i_l(v'_t)))\\
&- (1-\lambda)(\log D^i_s(G^i(v^i_s)) + \log(1 - D^i_s(v^i_t)))
\end{split}
\end{equation}
where $\lambda$ is a coefficient that we call global confidence, which balances the contributions of the two discriminators in updating the generator. In practice, we find that setting $\lambda$ to 0.5 for each subspace works well for the final result. 

Additionally, as the similarities between the entire distribution and the distribution of different subspaces are different, it is justified to use different lambdas for different subspaces instead of using a single one. We therefore propose a metric to set $\lambda$ dynamically, based on the proportion of the eigenvalue divergence between the two subspaces and the eigenvalue divergence between the whole source and target distributions, as shown in (\ref{equation_lambda}). In this paper, we only report results with dynamic $\lambda$. 
\begin{equation}
\label{equation_lambda}
\begin{split}
\lambda = \frac{EV\!D(V^i_s, V^i_t)}{EV\!D(V_s, V_t)}
\end{split}
\end{equation}

The eigenvalue divergence between two embedding distributions $V_1$ and $V_2$ can be computed as shown in (\ref{equation_evd}), where $e^{V_1}_k$ and $e^{V_2}_k$  represent the eigenvalues of $V_1$ and $V_2$. 
\begin{equation}
\label{equation_evd}
\begin{split}
EV\!D(V_1, V_2) = \sum_{k=1}^{d}(\log e^{V_1}_k - \log e^{V_2}_k)^2
\end{split}
\end{equation}

All subspace-specific generators are initialized with the single linear mapping discussed in Section~\ref{section_single_linear_mapping}.

\subsection{Subspace alignment}

The above multi-discriminator GAN assumes that we have an alignment between source subspaces and target subspaces.  We first present the method we use to produce aligned subspaces in both source and target distributions, and then the clustering method we use to find coherent subspaces, which are both important for the model's improved performance.

If we want to encourage words from a specific source subspace to be trained against words from a matching target subspace, we need to align the two cross-language subspaces. 
The second problem we need to solve for our multi-adversarial method to work is how to discover this alignment.
Although metrics such as Gromov-Hausdorff distance (GH) \cite{patra2019bilinguallexicon} and Eigenvalue Divergence (EVD) \cite{dubossarsky2020lostin} can be used to measure the similarity between two distributions and find the most similar target subspace for a given source subspace, matching between two sub-distributions may amplify any bias generated during the clustering. 

To avoid this problem, we only run the clustering on the source side. For a given source embedding space $V_s$, we denote its subspaces after clustering as $\left \{  V^1_s, V^2_s, ..., V^i_s, ..., V^n_s  \right \},$ where $n$ represent the number of subspaces. To align target words to their matching source subspace, we propose to first learn a single linear mapping from source to target space using the GAN-based method (without refinement) described previously, and then use the transpose of this linear mapping to retrieve the translation of each target word in the source language (using cross-domain similarity local scaling, defined below in Section~\ref{section_training}).  The subspace index of the target word is then set to the subspace index of this translation.  In this way, the target embedding space $V_t$ is partitioned into as many subspaces as the source embedding space, denoted as $\left \{ V^1_t, V^2_t, ..., V^i_t, ..., V^n_t \right \}$. This gives us aligned subspace pairs $(V^i_s, V^i_t)$.

Although the single linear mapping from source language to target language is not good enough to get accurate translations, our experiments indicate that it is a good method to produce a subspace alignment. A possible reason for this result is that the clustering on the source language has already grouped similar words. Therefore, even if a translation turns out to be incorrect, it usually has the same subspace index as the best translation.

\paragraph{Parameter-free hierarchical clustering}

A major issue in clustering an embedding space is how to find a clustering that adapts to the space, without fixed parameters.
To avoid having to identify the number of subspaces in advance, we use  hierarchical clustering. Recent work proposes a parameter-free method called First Integer Neighbor Clustering Hierarchy (FINCH) \cite{sarfraz2019efficientparameterfree}, which we use in this paper.  
Traditionally,  clustering methods split a given space of vectors into different clusters by calculating the distances between the centroid and the other vectors. FINCH is developed based on the observation that the first neighbour of each vector is a sufficient statistic to find links in the space, so that computing the distance matrix between all the vectors is not needed \cite{sarfraz2019efficientparameterfree}. For a given vector space, one first computes an adjacency link matrix using the  equation in (\ref{equation_clustering}).
\begin{equation}
\label{equation_clustering}
A(i, j) = \left\{\begin{matrix}
1 & i\!f~ j=\kappa^1_i ~or~ \kappa^1_j = i ~or~ \kappa^1_i=\kappa^1_j \\ 
0 & otherwise
\end{matrix}\right.
\end{equation}
where $i, j$ denote the indices of vectors and $\kappa^1_i$ represents the index of the first neighbour of the vector with index $i$. The connected components can then be detected from the adjacency matrix $A$ by building a directed or undirected graph on $A$. No parameter needs to be set. When the clustering on the original first level (original data) is completed, the centroid of each cluster can then be considered as a data vector for the next level and a new level of clustering is computed using the same procedure. In theory, all the vectors will eventually be gathered into a single cluster. In practice, we find that using the clusters of the last level or the second-to-last level  works well for our system.\footnote{In the code of \citet{sarfraz2019efficientparameterfree}, the last level means the level before grouping all the data vectors into a single cluster.}

\subsection{Training the GANs}
\label{section_training}


Training the GANs described in Sections~\ref{section_single_linear_mapping} and~\ref{section_multi_linear_mapping} can be challenging.  Based on previous work and our experience, we employ the following techniques during training.

\paragraph{Orthogonalization}

Previous work shows that enforcing the mapping matrix $W$ to be orthogonal during the training can improve the performance \cite{smith2017offlinebilingual}. In the system of \citet{conneau2018wordtranslation}, they follow the work of \citet{cisse2017parsevalnetworks} and approximate setting $W$ to an  orthogonal matrix with
$W\leftarrow (1+\beta)W -\beta( WW^\top)W$.
This orthogonalization usually performs well when setting $\beta$ to 0.001 \cite{conneau2018wordtranslation, wang2019weaklysupervised}.


\paragraph{Cross-Domain  Similarity Local  Scaling}

The trained mapping matrix $W$ can be used for retrieving the translation for a given source word $w_s$ by searching a target word $w_t$ whose embedding vector $v_t$ is close to $Wv_s$. But \citet{conneau2018wordtranslation} showed that using cross-domain similarity local scaling (CSLS) to retrieve translations is more accurate than standard nearest neighbor techniques and can reduce the impact of the hubs problem \cite{radovanovic2010hubsin, dinu2015Improvingzeroshot}.  Instead of just considering the distance between $Wv_s$ and $v_t$, CSLS also takes into account the neighbours of $v_t$ in the source language by minimising
$(
2\cos(Wv_s, v_t)-r_t(Wv_s)-r_s(v_t)
)$,
where $r_t(Wv_s)$ denotes the mean similarity between a $Wv_s$ and its neighbours in the target language, while $r_s(v_t)$ represents the mean similarity between $v_t$ and its neighbours in the source language.

\paragraph{Model selection criterion}
The cosine-based model selection criterion is another important component of adversarial training for selecting the best mapping matrix $W$. More specifically, at the end of each training epoch,  the current mapping is used to translate the ten-thousand most frequent source words into target words and calculate the average cosine similarity between the source vectors and the target vectors. This cosine-based criterion has been shown to correlate well with the quality of $W$ \cite{conneau2018wordtranslation, hartmann2019comparingunsupervised}.

\paragraph{Random restarts} 
Previous work \cite{vulic2019dowe,glavas2019howto} shows that using GANs to train the mapping matrix $W$ is not stable. \citet{hartmann2019comparingunsupervised} propose to solve this problem with the random restart technique. More specifically, before going to the step of refinement, they randomly train ten mapping matrices, choosing only the best model among them for the next step. The best model is selected with the unsupervised model selection criterion. Their experiments show that this model selection method has the best performance on bilingual lexicon induction. We follow \cite{vulic2019dowe,glavas2019howto} and apply the same random restart technique to train the single linear mapping and use it to initialize each subspace-specific generator.

\section{CLWE mapping refinement }
\label{section_refinement}

As in previous work, after GANs have been used to find a mapping from source to target word embeddings, a refinement step can be used to improve this mapping.  Refinement involves first inducing a  seed dictionary of word translations, and then refining the mapping using this seed dictionary.

\paragraph{Bidirectional seed dictionary induction}
Using the mapping learned with adversarial training, the translations $(w_{t_{1}}, w_{t_{2}},..., w_{t_{10000}})$ for the top ten-thousand source words $(w_{s_{1}}, w_{s_{2}},..., w_{s_{10000}})$ are retrieved and then back-translated into the source language $(w'_{s_{1}}, w'_{s_{2}},..., w'_{s_{10000}})$. The mutual translation pairs $(w_{s_{i}}, w_{t_{i}}) $ such that $w_{s_{i}} = w'_{s_{i}}$ constitute the seed dictionary. This guarantees that the induced seed dictionary will be bidirectional.

\paragraph{Mapping refinement}
The refinement step  is based on the Procrustes Analysis \cite{Schoneman1966ageneralized}. With the seed dictionary, the mapping can be updated using the objective in equation~\eqref{equation_mapping_mikolov}, and forced  to be orthogonal using singular value decomposition (SVD) \cite{xing2015normalizedword}.
%
%
Later work  combines the Procrustes Analysis with stochastic dictionary induction \cite{artetxe2018arobust} and greatly improves the performance of the standard refinement \cite{hartmann2019comparingunsupervised}. More specifically, in order to prevent local optima, after each iteration  some elements of the similarity matrix are randomly dropped, so that the similarity distributions of words change randomly and the new seed dictionary for the next iteration varies.

\paragraph{Global and local refinement}
Refinement can be applied to our multi-linear mapping in two different ways. First, after the training of all the subspace alignments, we can refine a linear relationship between the transformed source embeddings and the target embeddings, like previous unsupervised methods. This we call \textit{global refinement}. It is noteworthy that the combination of the multi-linear mapping trained by our multi-discriminator model and the refined single linear mapping is still multi-linear. Second, we can also refine the mapping of each subspace separately. More concretely, for a given subspace $(V^i_s,V^i_t)$, we build a local seed dictionary and use the local seed dictionary to update the mapping $G^i(v^i_s)$. We call this \textit{local refinement}. We evaluate both global and local refinement in the next section.

\section{Experiments}
\label{section_experiments}

Bilingual lexicon induction (BLI) has become a standard task for evaluating CLWE models. However, according to \citet{glavas2019howto} and \citet{mozhi2020whyoverfitting},  BLI performance of a given CLWE model doesn't always correlate with performance in other cross-lingual downstream tasks. In this section, we evaluate our proposal on both the task of BLI and the task of cross-lingual document classification (CLDC).

We evaluate our system both with and without refinement. Since GAN-based methods of learning CLWE are often criticized for their instability at inducing the seed dictionary, we report the average over 10 runs for the BLI without-refinement setting. We include the random restart technique for other tasks and report the result of the best model selected by the unsupervised model selection criterion. We evaluate our model both with global refinement (G-Ref) and local refinement (L-Ref). 
\vspace{-0.1cm}

\paragraph{BLI setting}  We use the dataset provided by \citet{conneau2018wordtranslation} for the task of BLI. This dataset contains high quality dictionaries for more than 150 language pairs. For each language pair, it provides a training dictionary of 5000 words and a test dictionary of 1500 words. This dataset allows us to have a better understanding of the performance of our proposal on many different language pairs. For each language pair, we retrieve the best translations of source words in the test dictionary using CSLS, and we report the accuracy with precision at one (P@1).
\vspace{-0.1cm}
 
 \paragraph{CLDC setting} We use the multilingual classification benchmark (MLDoc) provided by \citet{schwenk2018acorpus} for the task of CLDC. MLDoc contains training and test documents with balanced class priors for eight languages: German (de), English (en), Spanish (es), French (fr), Italian (it), Japanese (ja), Russian (ru) and Chinese (zh). We follow previous works \cite{glavas2019howto, mozhi2020whyoverfitting} and train a CNN classifier   on English using 10,000 documents and test the classifier on the other seven languages.\footnote{https://github.com/zhangmozhi/retrofit\_clwe} Each language contains 4000 test documents. The input of the classifier comes from the CLWE models. We report the average accuracy over ten runs. 
 \vspace{-0.1cm}
 
\paragraph{Language pairs} In this paper, we focus on projecting foreign language embeddings into the English space. We choose the eight languages included in MLDoc for both the BLI and CLDC tasks. Within the seven non-English languages, Japanese, Russian and Chinese are languages distant from English and the others are languages similar to English. For the task of BLI, we also investigate Turkish, another language distant from English.
\vspace{-0.1cm}

\paragraph{Monolingual word embeddings} We use the pretrained FastText embedding models \cite{bojanowski2017enrichingword} for our experiments. These embeddings of 300 dimensions are pretrained on Wikipedia dumps and publicly available.\footnote{https://fasttext.cc/docs/en/pretrained-vectors.html} Following previous works, we use the first 200,000 most frequent words for each monolingual embedding model.\footnote{The original pretrained Latvian fastText model only consists of 171,000 words.} We apply iterative normalization \cite{mozhi2019aregirls} on each embedding model before training. 

\vspace{-0.2cm}

\begin{table}
\setul{0.3ex}{0.7pt}
\addtolength{\tabcolsep}{-0.6ex}
\small
\begin{tabular}{@{}lrrrrrrrr} 
\toprule
\multicolumn{9}{c}{BLI Task - with refinement} \\ 
\midrule
 & \multicolumn{1}{c}{de} & \multicolumn{1}{c}{es} & \multicolumn{1}{c}{fr} & \multicolumn{1}{c}{it} & \multicolumn{1}{c}{ja} & \multicolumn{1}{c}{ru} & \multicolumn{1}{c}{tr} & \multicolumn{1}{c}{zh} \\ 
\midrule
PROC & 73.1 & 83.6 & 82.2 & 77.5 & 37.9 & 64.3 & \ul{63.1} & 40.0 \\
RCSLS & 73.1 & 83.1 & \ul{83.1} & \ul{78.9} & \ul{39.3} & \ul{64.6} & 63.1 & \ul{43.0} \\ 
\midrule
\midrule
MUSE & 73.7 & 83.0 & 82.2 & 78.5 & 29.3 & 62.7 & 60.5 & 38.1 \\
VecMap & 73.6 & \textbf{\ul{83.7}} & \textbf{82.9} & 78.5 & \textbf{34.7} & 63.1 & \textbf{61.3} & 36.4 \\ 
\midrule
Ours GRef & \textbf{\ul{74.1}} & \textbf{\ul{83.7}} & 82.4 & \textbf{78.6} & 34.1 & \textbf{64.0} & 61.2 & \textbf{38.2} \\
Ours LRef & 66.6 & 79.3 & 77.8 & 70.3 & 23.7 & 46.5 & 39.7 & 29.7 \\
\bottomrule
\end{tabular}\caption{BLI task results with refinement.  Bold shows the best score within unsupervised systems and underline shows the best score over all the systems.}
\label{table_bli_results_with_refinement}
\vspace{-0.3cm}

\end{table}

\begin{table}
\addtolength{\tabcolsep}{-0.5ex}
\centering
\small
\begin{tabular}{@{}lrrrrrrrr}
\toprule
\multicolumn{9}{c}{BLI Task - without refinement} \\ 
 & \multicolumn{1}{c}{de} & \multicolumn{1}{c}{es} & \multicolumn{1}{c}{fr} & \multicolumn{1}{c}{it} & \multicolumn{1}{c}{ja} & \multicolumn{1}{c}{ru} & \multicolumn{1}{c}{tr} & \multicolumn{1}{c}{zh}\\\midrule
 & \multicolumn{8}{c}{Successful runs averages}\\
MUSE   & 53.9 & 68.9 & 66.9 & \textbf{60.7}& 14.7 & \textbf{38.1} & 22.3 &  16.2   \\
VecMap & 0.0   & 0.0  &  0.0 & 0.0         & 0.0  & 0.0           & 0.0 &     0.0 \\ 
\midrule
Ours & \textbf{55.5} & \textbf{69.3} & \textbf{67.3} & 59.3 & \textbf{18.3} & \textbf{38.1} & \textbf{28.4} & \textbf{19.1} \\
\midrule
 & \multicolumn{8}{c}{Failures}     
\\ 
MUSE   &  3 &  1 & 0 &  1 &  5  & 5  & 4 & 9 \\
VecMap & 10 & 10 & 10& 10 & 10  & 10 & 10& 10\\ 
\midrule
Ours & 1 & 1 & 0 & 0 & 5& 4& 4 & 8\\
\bottomrule
\end{tabular}
\caption{BLI task results for unsupervised models without refinement. We consider accuracy below 2\% as failure and report the average accuracy with P@1 over the successful runs. Bold represents the best score.}
\label{table_bli_results_without_refinement}
\vspace{-0.5cm}
\end{table}

\paragraph{Baselines} The objective of our proposal is to improve the mapping ability of GANs by learning a multi-linear mapping instead of only a single-linear mapping. Therefore, we use the GAN-based system MUSE \citep{conneau2018wordtranslation}\footnote{https://github.com/facebookresearch/MUSE} as our main unsupervised baseline. Since the unsupervised method proposed by \citet{artetxe2018arobust}\footnote{ https://github.com/artetxem/vecmap} is considered a robust CLWE system, we also use it as our second unsupervised baseline (VecMap in the tables). In the setting with refinement, we use the iterative refinement with stochastic dictionary induction for all the unsupervised systems.\footnote{We disabled the re-weighting technique since it's not applicable for L-Ref. However, adding re-weighting to VecMap, MUSE and G-Ref doesn't change the gaps between them.} We also include two supervised systems, Procrustes (PROC) \cite{conneau2018wordtranslation} and Relaxed CSLS (RCSLS) \cite{armand2018lossin}, to better understand our method.  Both PROC and RCSLS are robust supervised systems for learning CLWE and have been widely used previously \cite{glavas2019howto, mozhi2020whyoverfitting}. We also wanted to include the supervised system proposed by \citet{nadapa2018normaneighborhood}, which learns multiple local mappings between embedding spaces, but their code is not publicly available.

\subsection{Performance on BLI}

We report the results of BLI both with and without refinement in Tables \ref{table_bli_results_with_refinement} and  \ref{table_bli_results_without_refinement}, respectively.

The results in Table \ref{table_bli_results_without_refinement} show a clear improvement from our multi-linear mapping, compared with single-linear GANs. We perform better than MUSE for both average accuracy and number of failures in almost every language pair. The advantages are more striking on distant language pairs.

VecMap is considered the most robust unsupervised model for learning CLWE. However, according to \citet{hartmann2019comparingunsupervised}, the advantage of VecMap mostly comes  from its refinement technique. From the results in Table \ref{table_bli_results_with_refinement}, we can see that when using the same refinement technique, our best model selected from ten random restarts using the unsupervised metric performs as well as VecMap or even better. Our model achieves higher scores on four language pairs and comes close for the other language pairs.

The results shown in Table \ref{table_bli_results_with_refinement} demonstrate that our model is comparable with supervised systems when using iterative refinement with stochastic dictionary induction and random restarts. We even perform better than PROC and RCSLS on similar language paris such as German and Spanish to English (de-en and es-en).

From the results in Table \ref{table_bli_results_with_refinement}, we can easily see that the global refinement outperforms local refinement. Using local refinement we even perform much worse than our GAN-based baseline. This phenomenon does not surprise us since local refinement can easily lead to overfitting on a given subspace, and we leave the investigation of alternative refinement methods to future work.

\subsection{Performance on CLDC}
We report the results on the task of CLDC without and with refinement in Tables \ref{table_cldc_results_without_refinement} and \ref{table_cldc_results_with_refinement}, respectively.

From the results shown in Table  \ref{table_cldc_results_without_refinement}, we can see that our multi-linear model continues to maintain its advantage over the  single-linear GAN, MUSE, in the setting without refinement. When refinement is added, MUSE becomes a little better than our model on German and Spanish. However, we still perform better on all those languages that are distant from English.

Differently from the task of BLI, there is no obvious advantage in the supervised baselines over our multi-linear model both with and without refinement. Conversely, as the results in Table \ref{table_cldc_results_without_refinement} indicate, our  model without refinement performs comparably or better than either our supervised baselines or VecMap in the setting with refinement. For example, our model achieves 69.5 of accuracy on Chinese test data, while the best supervised model, PROC, only has 32.0 accuracy. 

\begin{figure}
\centering
\includegraphics[scale=0.62]{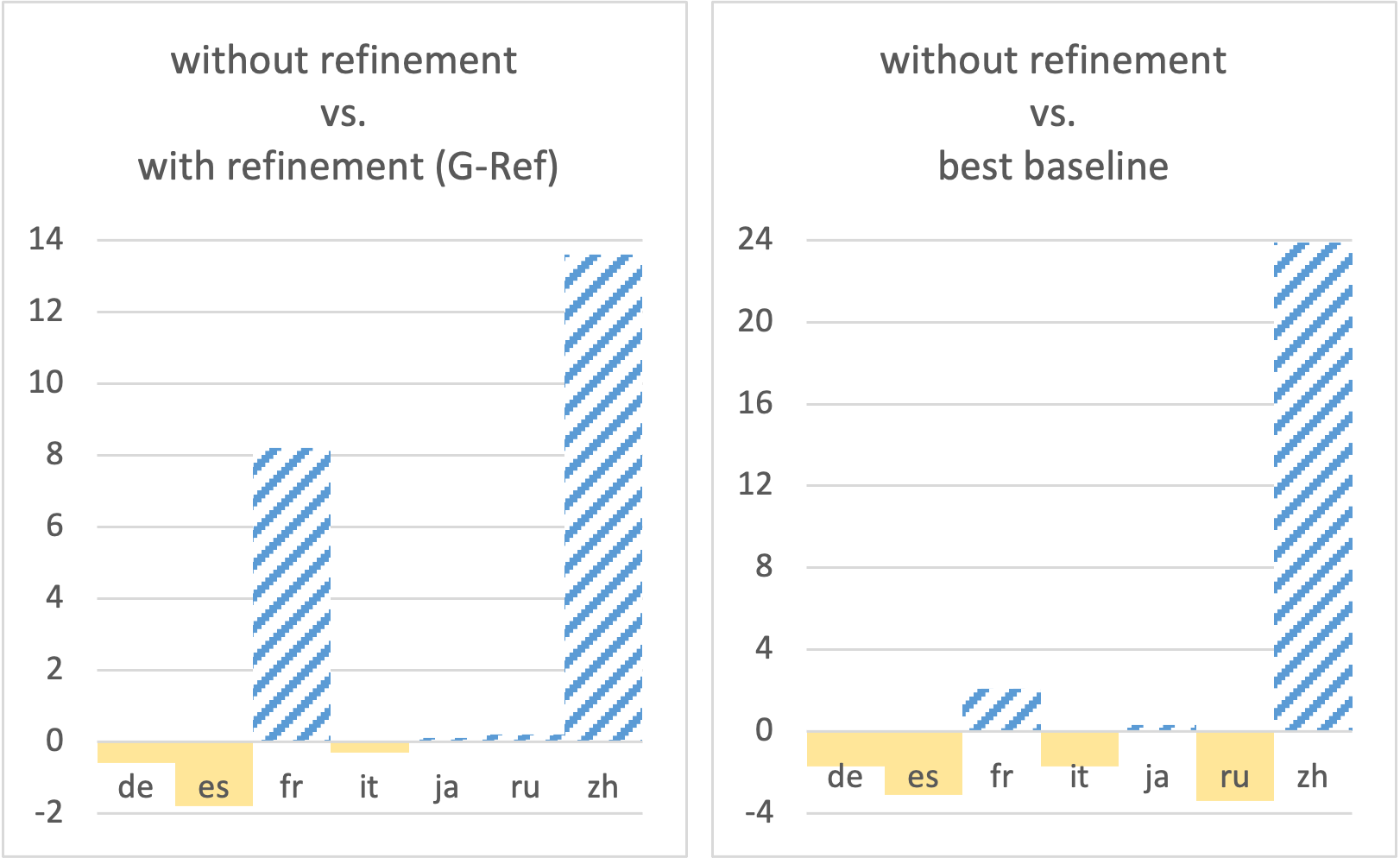}
\caption{Performance gap on the CLDC task. The left panel represents the gap between our multi-linear model with refinement and without refinement. The right panel represents the performance gap between our model without refinement and the best baseline (best results selected from our supervised and unsupervised baselines). Blue bars indicate the cases where the  model without refinement  performs better than its competitors. Yellow bars represent the opposite cases.}
\label{figure_cldc_gap}
\end{figure}

While CLWE refinement is a necessary step for the BLI task, for the CLDC task our model does not seem to need refinement. As the performance gap illustrated in Figure \ref{figure_cldc_gap} shows, our model performs worse when adding refinement for languages such as French, Japanese, Russian and Chinese, which includes all the languages distant from English.  Furthermore, even for languages where we benefit from refinement, the improvement is limited.

\begin{table}
\addtolength{\tabcolsep}{-0.6ex}
\centering
\begin{tabular}{@{}lrrrrrrrr} 
\toprule
\multicolumn{8}{c}{~CLDC Task - without refinement} \\ 
\midrule
 & \multicolumn{1}{c}{de} & \multicolumn{1}{c}{es} & \multicolumn{1}{c}{fr} & \multicolumn{1}{c}{it} & \multicolumn{1}{c}{ja} & \multicolumn{1}{c}{ru} & \multicolumn{1}{c}{zh} \\ 
\midrule
MUSE & 79.2 & \textbf{69.7} & 71.5 & 56.4 & 27.6 & 56.3 & 60.8 \\
VecMap & 21.4 & 24.6 & 23.1 & 23.7 & 21.2 & 23.6 & 22.3 \\ 
\hline
Ours & \textbf{80.1} & 67.4 & \textbf{73.2} & \textbf{62.3} & \textbf{30.3} & \textbf{61.5} & \textbf{69.5} \\
\bottomrule
\end{tabular}
\caption{CLDC results on MLDoc dataset \cite{schwenk2018acorpus} without refinement. Bold represents the best score.}
\label{table_cldc_results_without_refinement}
\vspace{-0.3cm}
\end{table}

\begin{table}
\setul{0.3ex}{0.7pt}
\addtolength{\tabcolsep}{-0.75ex}
\centering
\begin{tabular}{@{}lrrrrrrrr} 
\toprule
\multicolumn{8}{c}{CLDC Task - with refinement} \\ 
\midrule
 & \multicolumn{1}{c}{de} & \multicolumn{1}{c}{es} & \multicolumn{1}{c}{fr} & \multicolumn{1}{c}{it} & \multicolumn{1}{c}{ja} & \multicolumn{1}{c}{ru} & \multicolumn{1}{c}{zh} \\ 
\midrule
PROC & 81.4 & 69.6 & 70.7 & 62.9 & 30.0 & \ul{64.9} & 32.0 \\
RCSLS & 81.6 & \ul{70.5} & \ul{71.1} & 62.4 & 29.7 & 64.3 & 31.8 \\ 
\midrule
\midrule
MUSE & 80.9 & 69.4 & 70.6 & 59.9 & 28.9 & 61.1 & 45.6 \\
VecMap & \textbf{\ul{81.8}} & 69.8 & \textbf{71.0} & \textbf{\ul{64.0}} & 28.4 & 63.2 & 35.4 \\ 
\midrule
Ours G-Ref & 80.7 & 69.2 & 70.9 & 62.6 & \textbf{\ul{30.2}} & 61.3 & 55.9 \\
Ours L-Ref & 79.5 & 69.1 & 69.9 & 62.6 & 30.1 & 62.9 & \textbf{\ul{59.9}} \\
\bottomrule
\end{tabular}
\caption{CLDC results on MLDoc dataset \cite{schwenk2018acorpus} with refinement. Bold shows the best score within unsupervised systems and underline shows the best score over all the systems.}
\label{table_cldc_results_with_refinement}
\vspace{-0.3cm}
\end{table}

\section{Conclusion}

In this paper, we propose a multi-adversarial learning method for cross-lingual word embeddings. Our system learns different linear mappings for different source subspaces instead of just learning a single one for the whole source space. The results of our experiments on bilingual lexicon induction and cross-lingual document classification on both close languages and distant languages prove that learning cross-lingual word embeddings with a multi-linear mapping improves performance over a single-linear mapping. Future work will focus on learning multi-linear mappings for contextualized  embeddings.
 

\bibliography{multi_mapping}

\begin{thebibliography}{34}
\expandafter\ifx\csname natexlab\endcsname\relax\def\natexlab#1{#1}\fi

\bibitem[{Alvarez-Melis and
  Jaakkola(2018)}]{alvarezmelis2018gromovwassersteinalignment}
David Alvarez-Melis and Tommi~S. Jaakkola. 2018.
\newblock {Gromov-Wasserstein Alignment of Word Embedding Spaces}.
\newblock In \emph{Proceedings of the 2018 Conference on Empirical Methods in
  Natural Language Processing}, volume 1881-1890, Brussels, Belgium.

\bibitem[{Artetxe et~al.(2017)Artetxe, Labaka, and
  Agirre}]{artetxe2017learningbilingual}
Mikel Artetxe, Gorka Labaka, and Eneko Agirre. 2017.
\newblock {Learning bilingual word embeddings with (almost) no bilingual data}.
\newblock In \emph{Proceedings of the 55th Annual Meeting of the Association
  for Computational Linguistics}, pages 451--462, Vancouver, Canada.

\bibitem[{Artetxe et~al.(2018)Artetxe, Labaka, and Agirre}]{artetxe2018arobust}
Mikel Artetxe, Gorka Labaka, and Eneko Agirre. 2018.
\newblock {A robust self-learning method for fully unsupervised cross-lingual
  mappings of word embeddings}.
\newblock In \emph{Proceedings of the 56th Annual Meeting of the Association
  for Computational Linguistics}, volume Long Papers, pages 789--798,
  Melbourne, Australia.

\bibitem[{Barone(2016)}]{barone2016towardscrosslingual}
Antonio Valerio~Miceli Barone. 2016.
\newblock {Towards Cross-Lingual Distributed Representations without Parallel
  Text Trained with Adversarial Autoencoders}.
\newblock In \emph{Proceedings of the 1st Workshop on Representation Learning
  for NLP}, pages 121--126, Berlin, Germany.

\bibitem[{Bojanowski et~al.(2017)Bojanowski, Grave, Joulin, and
  Mikolov}]{bojanowski2017enrichingword}
Piotr Bojanowski, Edouard Grave, Armand Joulin, and Tomas Mikolov. 2017.
\newblock \href {https://doi.org/10.1162/tacl\_a\_00051} {{Enriching Word
  Vectors with Subword Information}}.
\newblock \emph{Transactions of the Association for Computational Linguistics},
  5:135--146.

\bibitem[{Cisse et~al.(2017)Cisse, Bojanowski, Grave, Dauphin, and
  Usunier}]{cisse2017parsevalnetworks}
Moustapha Cisse, Piotr Bojanowski, Edouard Grave, Yann Dauphin, and Nicolas
  Usunier. 2017.
\newblock {Parseval Networks: Improving Robustness to Adversarial Examples}.
\newblock In \emph{Proceedings of the 34th International Conference on Machine
  Learning}, volume~70, pages 854--863, Sydney, Australia.

\bibitem[{Conneau et~al.(2018)Conneau, Lample, Ranzato, Denoyer, and
  Jégou}]{conneau2018wordtranslation}
Alexis Conneau, Guillaume Lample, Marc'Aurelio Ranzato, Ludovic Denoyer, and
  Hervé Jégou. 2018.
\newblock {Word Translation Without Parallel Data}.
\newblock In \emph{Proceedings of the 6th International Conference on Learning
  Representations}, pages 1--14, Vancouver, Canada.

\bibitem[{Dinu et~al.(2015)Dinu, Lazaridou, and
  Baroni}]{dinu2015Improvingzeroshot}
Georgiana Dinu, Angeliki Lazaridou, and Marco Baroni. 2015.
\newblock {Improving Zero-Shot Learning by Mitigating the Hubness Problem}.
\newblock In \emph{Proceedings of the 3rd International Conference on Learning
  Representations}, volume Workshop Track, pages 1--10, Toulon, France.

\bibitem[{Dubossarsky et~al.(2020)Dubossarsky, Vulić, Reichart, and
  Korhonen}]{dubossarsky2020lostin}
Haim Dubossarsky, Ivan Vulić, Roi Reichart, and Anna Korhonen. 2020.
\newblock \href {http://arxiv.org/abs/2001.11136} {{Lost in Embedding Space:
  Explaining Cross-Lingual Task Performance with Eigenvalue Divergence}}.
\newblock \emph{arXiv}, pages 1--10.

\bibitem[{Glavaš et~al.(2019)Glavaš, Litschko, Ruder, and
  Vulić}]{glavas2019howto}
Goran Glavaš, Robert Litschko, Sebastian Ruder, and Ivan Vulić. 2019.
\newblock {How to (Properly) Evaluate Cross-Lingual Word Embeddings: On Strong
  Baselines, Comparative Analyses, and Some Misconceptions}.
\newblock In \emph{Proceedings of the 57th Annual Meeting of the Association
  for Computational Linguistics}, pages 710--721, Florence, Italy.

\bibitem[{Goodfellow et~al.(2014)Goodfellow, Pouget-Abadie, Mirza, Xu,
  Warde-Farley, Ozair, Courville, and
  Bengio}]{goodfellow2014generativeadversarial}
Ian~J. Goodfellow, Jean Pouget-Abadie, Mehdi Mirza, Bing Xu, David
  Warde-Farley, Sherjil Ozair, Aaron Courville, and Yoshua Bengio. 2014.
\newblock {Generative Adversarial Nets}.
\newblock In \emph{Proceedings of the 27th International Conference on Neural
  Information Processing Systems}, page 2672–2680, Montréal, Canada.

\bibitem[{Guo et~al.(2015)Guo, Che, Yarowsky, Wang, and
  Liu}]{guo2015crosslingualdependency}
Jiang Guo, Wanxiang Che, David Yarowsky, Haifeng Wang, and Ting Liu. 2015.
\newblock \href {https://doi.org/10.3115/v1/p15-1119} {{Cross-lingual
  Dependency Parsing Based on Distributed Representations}}.
\newblock In \emph{Proceedings of the 53rd Annual Meeting of the Association
  for Computational Linguistics and the 7th International Joint Conference on
  Natural Language Processing}, pages 1234--1244, Beijing, China.

\bibitem[{Hartmann et~al.(2019)Hartmann, Kementchedjhieva, and
  Søgaard}]{hartmann2019comparingunsupervised}
Mareike Hartmann, Yova Kementchedjhieva, and Anders Søgaard. 2019.
\newblock {Comparing Unsupervised Word Translation Methods Step by Step}.
\newblock In \emph{Proceedings of the 33rd Conference on Neural Information
  Processing Systems}, pages 6033--6043, Vancouver, Canada.

\bibitem[{Hoshen and Wolf(2018)}]{hoshen2018aniterative}
Yedid Hoshen and Lior Wolf. 2018.
\newblock \href {http://arxiv.org/abs/1801.06126v1} {{An Iterative Closest
  Point Method for Unsupervised Word Translation}}.
\newblock \emph{ArXiv}.

\bibitem[{Joulin et~al.(2018)Joulin, Bojanowski, Mikolov, Jegou, and
  Grave}]{armand2018lossin}
Armand Joulin, Piotr Bojanowski, Tomas Mikolov, Herve Jegou, and Edouard Grave.
  2018.
\newblock {Loss in Translation: Learning Bilingual Word Mapping with a
  Retrieval Criterion}.
\newblock In \emph{Proceedings of the 2018 Conference on Empirical Methods in
  Natural Language Processing}, pages 2979--2984, Brussels, Belgium.

\bibitem[{Klementiev et~al.(2012)Klementiev, Titov, and
  Bhattarai}]{klementiev2012inducingcrosslingual}
Alexandre Klementiev, Ivan Titov, and Binod Bhattarai. 2012.
\newblock {Inducing Crosslingual Distributed Representations of Words}.
\newblock In \emph{Proceedings of the 24th International Conference on
  Computational Linguistics}, pages 1459--1473, Mumbai, India.

\bibitem[{Mikolov et~al.(2013)Mikolov, Chen, Corrado, and
  Dean}]{mikolov2013efficientestimation}
Tomas Mikolov, Kai Chen, Greg Corrado, and Jeffrey Dean. 2013.
\newblock {Efficient Estimation of Word Representations in Vector Space}.
\newblock In \emph{Proceedings of the 1st International Conference on Learning
  Representations}, pages 1--12, Arizona, USA.

\bibitem[{Nakashole(2018)}]{nadapa2018normaneighborhood}
Ndapa Nakashole. 2018.
\newblock \href {https://doi.org/10.18653/v1/d18-1047} {{NORMA: Neighborhood
  Sensitive Maps for Multilingual Word Embeddings}}.
\newblock In \emph{Proceedings of the 2018 Conference on Empirical Methods in
  Natural Language Processing}, pages 512--522, Brussels, Belgium.

\bibitem[{Patra et~al.(2019)Patra, Moniz, Garg, Gormley, and
  Neubig}]{patra2019bilinguallexicon}
Barun Patra, Joel Ruben~Antony Moniz, Sarthak Garg, Matthew~R. Gormley, and
  Graham Neubig. 2019.
\newblock {Bilingual Lexicon Induction with Semi-supervisionin Non-Isometric
  Embedding Spaces}.
\newblock In \emph{Proceedings of the 57th Annual Meeting of the Association
  for Computational Linguistic}, pages 184--193, Florence, Italy.

\bibitem[{Radovanović et~al.(2010)Radovanović, Nanopoulos, and
  Ivanović}]{radovanovic2010hubsin}
Milos Radovanović, Alexandros Nanopoulos, and Mirjana Ivanović. 2010.
\newblock {Hubs in Space: Popular Nearest Neighbors in High-Dimensional Data}.
\newblock \emph{Journal of Machine Learning Research}, 11:2487--2531.

\bibitem[{Ruder et~al.(2019)Ruder, Vulić, and Søgaard}]{ruder2019asurvey}
Sebastian Ruder, Ivan Vulić, and Anders Søgaard. 2019.
\newblock {A Survey Of Cross-lingual Word Embedding Models}.
\newblock \emph{Journal of Artificial Intelligence Research}, 65:569--631.

\bibitem[{Sarfraz et~al.(2019)Sarfraz, Sharma, and
  Stiefelhagen}]{sarfraz2019efficientparameterfree}
M.~Saquib Sarfraz, Vivek Sharma, and Rainer Stiefelhagen. 2019.
\newblock {Efficient Parameter-free Clustering Using First Neighbor Relations}.
\newblock In \emph{Proceedings of the 2019 IEEE Conference on Computer Vision
  and Pattern Recognition (CVPR)}, pages 8934--8943, Long Beach, California,
  USA.

\bibitem[{Schwenk and Li(2018)}]{schwenk2018acorpus}
Holger Schwenk and Xian Li. 2018.
\newblock {A Corpus for Multilingual Document Classification in Eight
  Languages}.
\newblock In \emph{Proceedings of the Eleventh International Conference on
  Language Resources and Evaluation}, Miyazaki, Japan.

\bibitem[{Schönemann(1966)}]{Schoneman1966ageneralized}
Peter~H. Schönemann. 1966.
\newblock {A generalized solution of the {Orthogonal Procrustes} problem}.
\newblock \emph{Psychometrika}, 31:1--10.

\bibitem[{Smith et~al.(2017)Smith, Turban, Hamblin, and
  Hammerla}]{smith2017offlinebilingual}
Samuel~L. Smith, David H.~P. Turban, Steven Hamblin, and Nils~Y. Hammerla.
  2017.
\newblock {Offline Bilingualword Vectors, Orthogonal Transformations and the
  Inverted Softmax}.
\newblock In \emph{Proceedings of the 15th International Conference on Learning
  Representations}, pages 1--10, Toulon, France.

\bibitem[{Søgaard et~al.(2018)Søgaard, Ruder, and Vulić}]{sogaard2018onthe}
Anders Søgaard, Sebastian Ruder, and Ivan Vulić. 2018.
\newblock {On the Limitations of Unsupervised Bilingual Dictionary Induction}.
\newblock In \emph{Proceedings of the 56th Annual Meeting of the Association
  for Computational Linguistics}, volume Long Papers, pages 778--788,
  Melbourne, Australia.

\bibitem[{Vulić et~al.(2019)Vulić, Glavaš, Reichart, and
  Korhonen}]{vulic2019dowe}
Ivan Vulić, Goran Glavaš, Roi Reichart, and Anna Korhonen. 2019.
\newblock {Do We Really Need Fully Unsupervised Cross-Lingual Embeddings?}
\newblock In \emph{Proceedings of the 2019 Conference on Empirical Methods in
  Natural Language Processing}, pages 4407--4418, Hong Kong, China.

\bibitem[{Vulić et~al.(2020)Vulić, Ruder, and Søgaard}]{vulic2020areall}
Ivan Vulić, Sebastian Ruder, and Anders Søgaard. 2020.
\newblock \href {http://arxiv.org/abs/2004.04070} {{Are All Good Word Vector
  Spaces Isomorphic?}}
\newblock \emph{arXiv}, pages 1--11.

\bibitem[{Wang et~al.(2019)Wang, Henderson, and
  Merlo}]{wang2019weaklysupervised}
Haozhou Wang, James Henderson, and Paola Merlo. 2019.
\newblock {Weakly-Supervised Concept-based Adversarial Learning for
  Cross-lingual Word Embeddings}.
\newblock In \emph{Proceedings of the 2019 Conference on Empirical Methods in
  Natural Language Processing}, pages 4419--4430, Hong Kong, China.

\bibitem[{Xing et~al.(2015)Xing, Wang, Liu, and Lin}]{xing2015normalizedword}
Chao Xing, Dong Wang, Chao Liu, and Yiye Lin. 2015.
\newblock {Normalized Word Embedding and Orthogonal Transform for Bilingual
  Word Translation}.
\newblock In \emph{Proceedings of the 2015 Conference of the North American
  Chapter of the Association for Computational Linguistics}, pages 1006--1011,
  Denver, Colorado.

\bibitem[{Zhang et~al.(2017)Zhang, Liu, Luan, and
  Sun}]{zhang2017adversarialtraining}
Meng Zhang, Yang Liu, Huanbo Luan, and Maosong Sun. 2017.
\newblock {Adversarial Training for Unsupervised Bilingual Lexicon Induction}.
\newblock In \emph{Proceedings of the 55th Annual Meeting of the Association
  for Computational Linguistics}, pages 1959--1970, Vancouver, Canada.

\bibitem[{Zhang et~al.(2020)Zhang, Fujinuma, Paul, and
  Boyd-Graber}]{mozhi2020whyoverfitting}
Mozhi Zhang, Yoshinari Fujinuma, Michael~J Paul, and Jordan Boyd-Graber. 2020.
\newblock \href {https://doi.org/10.18653/v1/2020.acl-main.201} {{Why
  Overfitting Isn’t Always Bad: Retrofitting Cross-Lingual Word Embeddings to
  Dictionaries}}.
\newblock In \emph{Proceedings of the 58th Annual Meeting of the Association
  for Computational Linguistics}, pages 2214--2220, Online.

\bibitem[{Zhang et~al.(2019)Zhang, Xu, Kawarabayashi, Jegelka, and
  Boyd-Graber}]{mozhi2019aregirls}
Mozhi Zhang, Keyulu Xu, Ken-ichi Kawarabayashi, Stefanie Jegelka, and Jordan
  Boyd-Graber. 2019.
\newblock \href {https://doi.org/10.18653/v1/p19-1307} {{Are Girls Neko or
  Shōjo? Cross-Lingual Alignment of Non-Isomorphic Embeddings with Iterative
  Normalization}}.
\newblock In \emph{Proceedings of the 57th Annual Meeting of the Association
  for Computational Linguistics}, pages 3180--3189.

\bibitem[{Zou et~al.(2013)Zou, Socher, Cer, and Manning}]{zou2013bilingualword}
Will~Y. Zou, Richard Socher, Daniel Cer, and Christopher~D. Manning. 2013.
\newblock {Bilingual Word Embeddings for Phrase-Based Machine Translation}.
\newblock In \emph{Proceedings of the 2013 Conference on Empirical Methods in
  Natural Language Processing}, pages 1393--1398, Seattle, Washington, USA.

\end{thebibliography}
\bibliographystyle{acl_natbib}

\end{document}